\documentclass[letterpaper, 10 pt, journal, twoside]{IEEEtran}

\usepackage{graphicx}
\usepackage{algorithm}
\usepackage{algpseudocode}
\usepackage{balance}
\usepackage[verbose]{wrapfig}
\usepackage[caption=false]{subfig}
\usepackage{multirow}
\usepackage[normalem]{ulem}
\input{terms.tex}

\newcommand\CONDITION[2]%
  {\begin{tabular}[t]{@{}l@{}l@{}}
     #1&#2
  \end{tabular}%
  }
\algdef{SE}[WHILE]{While}{EndWhile}[1]%
  {\algorithmicwhile\ \CONDITION{#1}{\ \algorithmicdo}}%
  {\algorithmicend\ \algorithmicwhile}
\algdef{SE}[FOR]{For}{EndFor}[1]%
  {\algorithmicfor\ \CONDITION{#1}{\ \algorithmicdo}}%
  {\algorithmicend\ \algorithmicfor}
\algdef{S}[FOR]{ForAll}[1]%
  {\algorithmicforall\ \CONDITION{#1}{\ \algorithmicdo}}
\algdef{SE}[REPEAT]{Repeat}{Until}{\algorithmicrepeat}[1]%
  {\algorithmicuntil\ \CONDITION{#1}{}}
\algdef{SE}[IF]{If}{EndIf}[1]%
  {\algorithmicif\ \CONDITION{#1}{\ \algorithmicthen}}%
  {\algorithmicend\ \algorithmicif}%
\algdef{C}[IF]{IF}{ElsIf}[1]%
  {\algorithmicelse\ \algorithmicif\ \CONDITION{#1}{\ \algorithmicthen}}

\title{Learning When to Trust a Dynamics Model\\for Planning in Reduced State Spaces}

\author{\dale$^{1}$, \tom$^{1}$, \peter$^{1}$ and \dmitry$^{1}$
\thanks{Manuscript received: September 11, 2019; Revised December 7, 2019; Accepted January 20, 2020.}
\thanks{This paper was recommended for publication by Editor Nancy Amato upon evaluation of the Associate Editor and Reviewers' comments. This work was supported in part by NSF grant IIS-1750489, ONR grant N000141712050, and by Toyota  Research  Institute  (TRI).  This  article  solely  reflects  the  opinions of  its  authors  and  not  TRI  or  any  other  Toyota  entity.}
\thanks{$^{1}$ All authors are with the University of Michigan Robotics Institute, Ann Arbor, MI, USA. {\tt\small \{dmcconac, tpower, pmitrano, dmitryb\}@umich.edu}}
\thanks{Digital Object Identifier (DOI): see top of this page.}
}

\begin{document}
\maketitle


\begin{abstract}
When the dynamics of a system are difficult to model and/or time-consuming to evaluate, such as in deformable object manipulation tasks, motion planning algorithms struggle to find feasible plans efficiently. Such problems are often reduced to state spaces where the dynamics are straightforward to model and evaluate. However, such reductions usually discard information about the system for the benefit of computational efficiency, leading to cases where the true and reduced dynamics disagree on the result of an action. This paper presents a formulation for planning in reduced state spaces that uses a classifier to bias the planner away from state-action pairs that are not reliably feasible under the true dynamics. We present a method to generate and label data to train such a classifier, as well as an application of our framework to rope manipulation, where we use a Virtual Elastic Band (VEB) approximation to the true dynamics. Our experiments with rope manipulation demonstrate that the classifier significantly improves the success rate of our RRT-based planner in several difficult scenarios which are designed to cause the VEB to produce incorrect predictions in key parts of the environment.
\end{abstract}



\vspace{-0.04in} 
\section{INTRODUCTION}
\vspace{-0.04in} 
\IEEEPARstart{R}{obot} motion planning algorithms have been extremely successful for systems where the dynamics can be easily specified and efficiently evaluated. However, for tasks such as manipulation of deformable objects, the dynamics are very difficult to model \cite{Essahbi2012} and usually require numerical simulation to evaluate. This simulation can be time-consuming and/or inaccurate. Including such simulations inside a planner can result in plans that take hours to compute \cite{Bai2016}.

Motivated by tasks where dynamics are difficult to specify and evaluate, we present a framework to plan in a reduced state space with simplified dynamics while biasing the planner to find plans that are likely to be feasible under the true dynamics. To do this, we define a function that maps from the true state space to a reduced state space as well as a dynamics model in the reduced space. We can then generate plans in the reduced space and roll them out on the true system offline to gather data on how the reduced and true dynamics correspond. Specifically, we find which transitions (i.e. state-action pairs) in the reduced state space produce reliable predictions as compared to rolling out the given action with the true dynamics. 

\begin{figure}[t]
    \centering
    \includegraphics[trim={0 1.5in 5.3in 0},clip,width=\columnwidth]{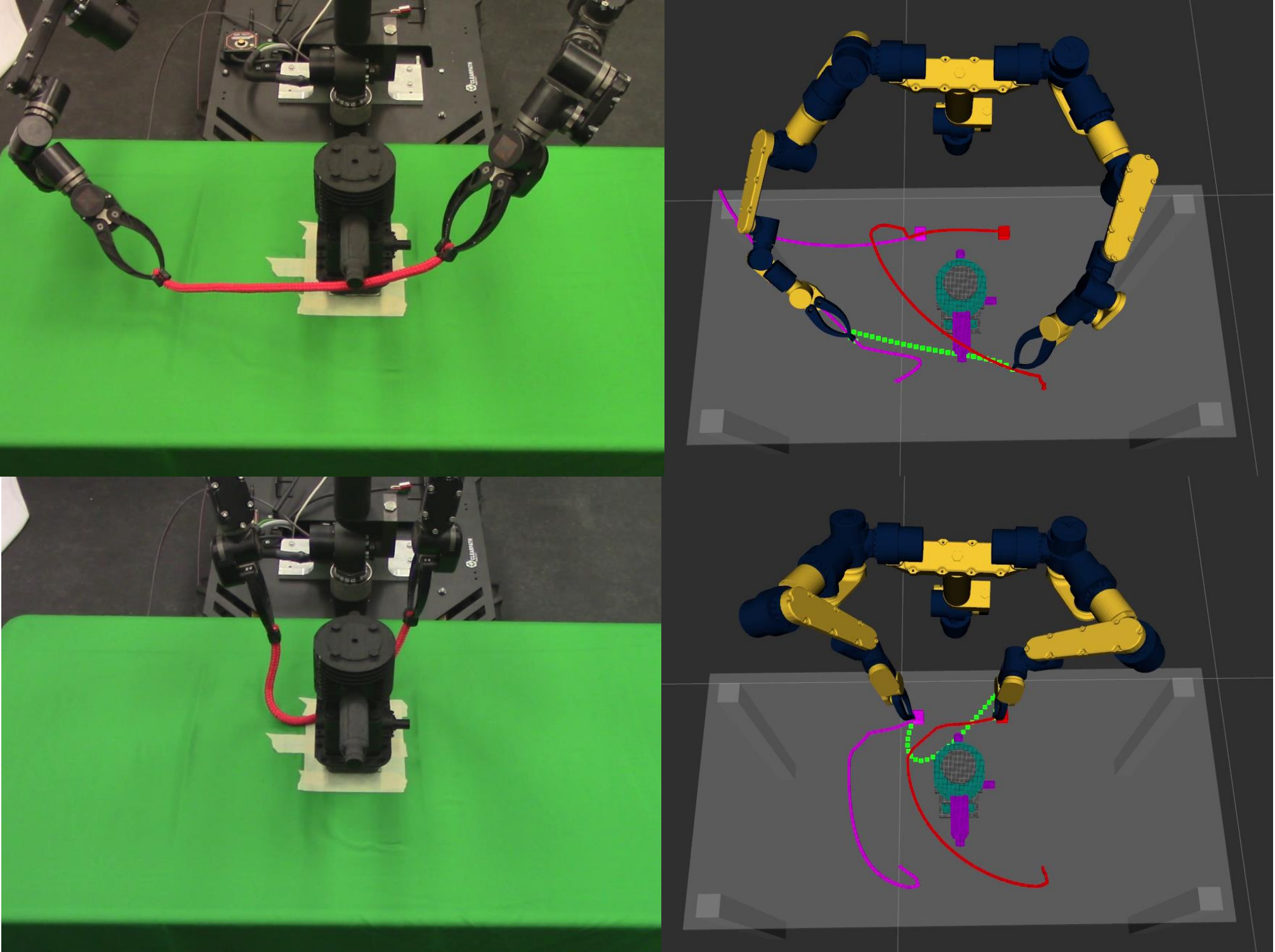}
    \caption{Top: a plan generated without using a classifier moves the rope under a hook and gets caught. Bottom: a plan generated with a classifier avoids this mistake, and reaches the goal.}
    \label{fig:intro_figure}
    \vspace{-0.15in}
\end{figure}

After gathering a dataset where transitions are labeled as reliable or unreliable, we train a classifier to predict the reliability of a given transition. We then incorporate this classifier into an RRT-based planner by biasing the planner to reject transitions that are classified as unreliable. The resulting planner tends to find sequences of transitions that are likely to produce the desired outcome when the true dynamics of the system are applied, even though the planner plans with no explicit knowledge of the true dynamics.

This paper presents both an abstract formulation of the problem of planning in a reduced state space with a classifier and how to apply this formulation to two systems. First, to illustrate our framework on a straightforward example, we consider a planar three-link arm with limited joint torque. Using a learned classifier for this system allows us to plan in configuration space (not considering dynamics) while avoiding transitions that cannot be accomplished with the limited torque. The second system focuses on rope manipulation tasks; we use a Virtual Elastic Band (VEB) \cite{McConachie2017} as the reduced dynamics model of the rope, as this has been shown to allow fast planning in difficult scenarios. However, this model assumes that there is no minimum length of the rope, which entails that the planner cannot detect cases where the slack material is caught on corners or hooks, preventing the motion plan from being completed because the caught object can overstretch (see Figure \ref{fig:intro_figure}). Thus, we learn a classifier to bias the planner away from states where the object can be caught on an obstacle.
The contributions of this paper are:
\begin{enumerate}
    \item A novel formulation of planning in reduced state spaces
    \item A method to generate and label data for classification of transition reliability
    \item Experiments demonstrating the efficacy of our framework for both the planar arm reaching and rope manipulation tasks
\end{enumerate}

Our experiments suggest that we can learn a classification function for the reliability of transitions which improves the success rate of planning with the reduced model for both the planar arm and rope manipulation. Our simulation experiments considering rope manipulation in several challenging environments containing hooks demonstrate the classifier's ability to bias the planner away from unreliable transitions and to generalize over environments and rope lengths. 
Finally, to show a practical application of our work, we demonstrate our method running on a 16 DoF robot manipulating a rope near an engine assembly.


\vspace{-0.04in} 
\section{RELATED WORK}
\vspace{-0.04in} 

Robotic manipulation of deformable objects has been studied in many contexts ranging from surgery to industrial manipulation (see \cite{Khalil2010} and \cite{Sanchez2018deformablesurvey} for extensive surveys).

Much work in deformable object manipulation relies on simulating an accurate model of the object being manipulated. 
The most common simulation methods use Mass-Spring models \cite{Gibson1997, Essahbi2012}, which are generally not accurate for large deformations \cite{Maris2010}, and Finite-Element (FEM) models \cite{Muller2002,Irving2004,Kaufmann2008}. FEM-based methods are widely used and physically well-founded, but they can be unstable when subject to contact constraints, which are especially important in this work. 

Motion planning for manipulation of deformable objects is an active area of research \cite{Jimenez2012} with many sampling-based planners proposed \cite{BurchanBayazit2002,Gayle2005,Moll2006,Saha2008,Roussel2015}. 
However, all the above methods either disallow contact with the environment or rely on potentially time-consuming physical simulation of the deformable object, which is often very sensitive to physical and computational parameters that may be difficult to determine. In contrast our method builds on \cite{McConachie2019}, which uses reduced models for motion planning with far lower computational cost.

In terms of applying machine learning to control and planning, prior work has primarily used learned dynamics models for control \cite{Jia2018,Finn2017,Banijamali2017,Zhang2019, Sutanto2019}. Recent work \cite{ichter2019} has also explored planning in learned reduced space, but they do not consider the error in a reduced model's prediction when planning. Visual Planning and Acting (VPA)~\cite{vpa2019rss} learns a latent transition model based on visual input for planning. This work also uses on a classifier to prune infeasible transitions during planning. However, despite this classifier, only 15\% of generated plans were visually plausible, with only 20\% of the visually plausible plans being executable. In this paper we do not focus on learning a reduction but rather on creating a framework that can be used to overcome limitations in a given model reduction. 


\vspace{-0.04in} 
\section{General Problem Formulation}
\vspace{-0.04in} 

We begin by formulating our problem in a system-agnostic way and then describe how to apply this formulation to planning for rope manipulation. Let the true system operate in a state space $\truestatespace$ with dynamics $\truestate_{t+1} = f(\truestate_t,\statecommand,\obstacle)$, where $\statecommand$ is a command given to the system and $\obstacle$ is the environment. We assume that the true state space has Markovian dynamics. 

The problem we address in this work is how to find a sequence of $\taskexecutiontime$ commands $\{\statecommand_1, \dots, \statecommand_{\taskexecutiontime} \}$ to move from a start state $\truestate_0$ to a goal state. The goal set is specified by the function $\texttt{Goal}_\truestate: \truestatespace \rightarrow \{ 0,1 \}$, which returns $1$ if a state is a goal and $0$ otherwise. We thus to seek to solve the following:

\begin{equation}
    \begin{aligned}
        & \text{find}   & & \taskexecutiontime, \{\statecommand_1, \dots, \statecommand_{\taskexecutiontime} \} \\
        & \text{s.t.}   & & \texttt{Goal}_\truestate(\truestate_{\taskexecutiontime}) = 1\\ 
        &               & & \truestate_t = f(\truestate_{t-1}, \statecommand_t,\obstacle), \:  i = 1, \dots, \taskexecutiontime\\
    \end{aligned}
    \label{eqn:main_planning_problem}
\end{equation}

\noindent However, $f$ may not be known in closed-form or it may be expensive to evaluate within a planner. Thus we cannot solve this problem by planning in $\truestatespace$ with the true dynamics.


To create a more tractable planning problem we define $\reducedstatespace$ to be a \textit{reduced state space} and define a reduction function: $\reducedstate = r(\truestate,\obstacle)$. We do not assume that $r$ is invertible. Dynamics in $\reducedstatespace$ are defined as $b_{t+1} = g(b_t,u^\reducedstate,\obstacle)$ (note that $u^\reducedstate$ and $\statecommand$ may be in different spaces). We treat the dynamics in this reduced state space as Markovian. 
$\reducedstatespace$, $r$, and $g$ are user-defined. There is then an analogous goal function for reduced states $\texttt{Goal}_\reducedstate: \reducedstatespace \rightarrow \{ 0,1 \}$.  The planning problem then becomes:

\begin{equation}
    \begin{aligned}
        & \text{find}   & & \taskexecutiontime, \{\reducedcommand_1, \dots, \reducedcommand_{\taskexecutiontime} \} \\
        & \text{s.t.}   & & \reducedstate_0 = r(\truestate_0,\obstacle) \\
        &               & &  \texttt{Goal}_\reducedstate(\reducedstate_{\taskexecutiontime}) = 1\\
        &               & & \reducedstate_{t} = g(\reducedstate_{t-1},\reducedcommand_t,\obstacle), \:  i = 1, \dots, \taskexecutiontime \\
    \end{aligned}
    \label{eqn:approx_planning_problem}
\end{equation}

Rather than making explicit guarantees on the relationship between $f$ and $g$, we assume we have access to a rollout function $\truestate_{t+1} = \Gamma(\truestate_t,u^\reducedstate,\obstacle)$, which outputs the next state when attempting to perform an action $u^\reducedstate$ given some controller for the system. We assume that $\Gamma$ has built-in safety limits, so it will stop before violating a constraint (e.g. stopping before colliding with an obstacle). If $\Gamma$ reaches a constraint boundary it will output the state on the boundary and will not violate the constraint. $\Gamma$ is treated as a black box. The form of $\Gamma$ may be known but even if it is, we assume it is too expensive to evaluate within the planner, otherwise there would likely be no need for the reduction. We assume we are able to gather data by executing $\Gamma$.

We will solve the problem in Equation \ref{eqn:approx_planning_problem} using a motion planner that plans in $\reducedstatespace$. However, the plan we generate may not lead to the goal in execution because we may have lost information in $r$ and/or $g$. Our approach is thus to bias our planner so that it avoids taking actions for which $r$ and $g$ are not reliable approximations of the behavior of the system. See Fig. \ref{fig:overview} for an overview of our framework.


\vspace{-0.04in} 
\section{Learning Transition Reliability}
\vspace{-0.04in} 

 To bias the planner that plans in $\reducedstatespace$ we will learn a classifier that attempts to predict if a given transition $T^\reducedstate = \left<b_t,u^\reducedstate,\obstacle\right>$ will reliably succeed (e.g. not be stopped by a constraint boundary) when executed in environment $\obstacle$. We thus wish to learn a function 
$\texttt{Classify}:\{T^\reducedstate\} \rightarrow \left\{\texttt{Reliable},\texttt{Unreliable}\right\}$, which outputs $\texttt{Reliable}$ when performing this transition reliably succeeds under various starting $\truestate_0$s and previous command sequences $u^b_{0:t-1}$, and $\texttt{Unreliable}$ otherwise, in which case the planner should be biased not to use $T^\reducedstate$.

Ideally, we would include $\truestate_0$ and $u^b_{0:t-1}$ as input to the classifier. However, $\truestatespace$ may be arbitrarily high-dimensional (e.g. for a deformable object) and there may be an arbitrary number of previous commands before $T^\reducedstate$, thus making the classifier very difficult to learn with a realistically-sized dataset. Instead we only consider $T^\reducedstate$ as input.

\vspace{-0.04in} 
\subsection{Data Generation and Labeling}
\vspace{-0.04in} 

\begin{figure}[t]
    \centering
    \includegraphics[width=\columnwidth]{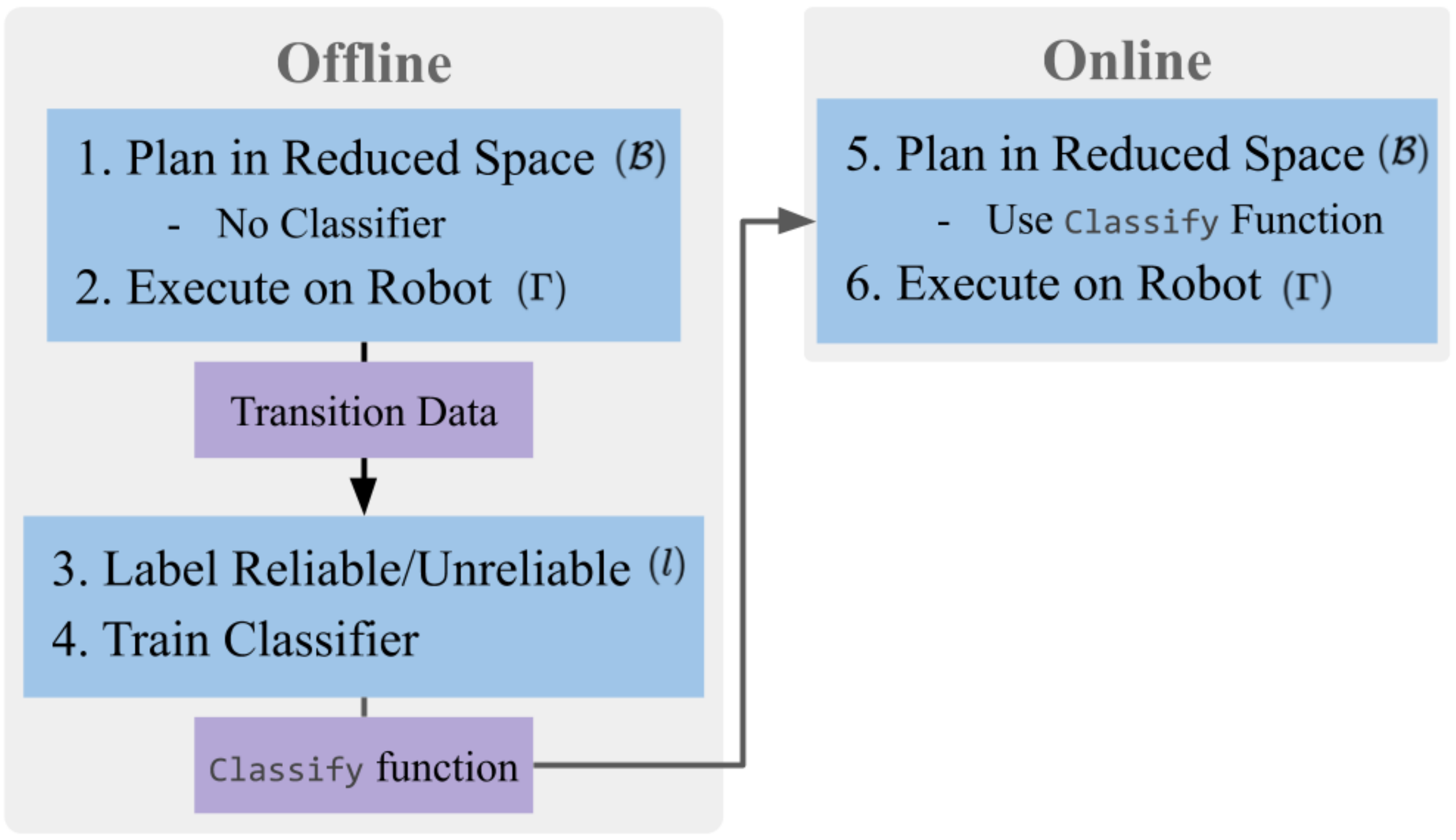}
    \vspace{-0.15in}
    \caption{An outline of our framework. First, we plan and execute many control sequences to gather a dataset of transitions. These transitions are labeled according to a function $l$ and used to train a classifier which predicts whether a transition is reliable given the model reduction. This classifier is used to bias the planner away from unreliable transitions.}
    \label{fig:overview}
    \vspace{0.1in}
    \centering
    \includegraphics[width=\linewidth]{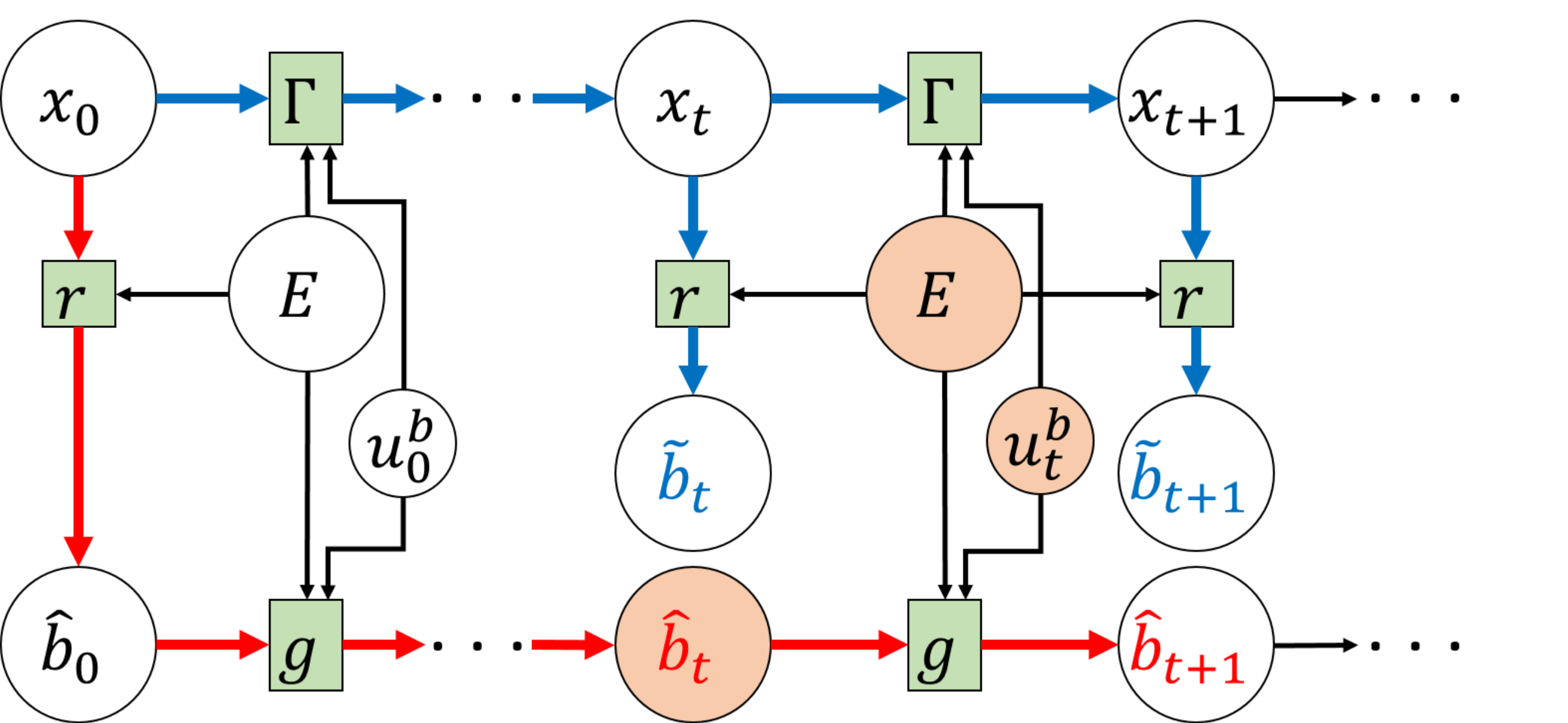}
    \vspace{-0.1in}
    \caption{Circles represent variables and boxes represent functions. Orange: variables defining the $t$th transition. Red path: reduced dynamics prediction; Blue path: rollout result.}
    \vspace{-0.1in}
    \label{fig:flowgraph}
\end{figure}

To train our classifier, we need to gather a dataset of transitions and label them by whether the model reduction produces a reliable prediction. We would like to generate a training dataset of transitions in a similar way to how a planner would generate transitions, since this avoids distribution mismatch problems when planning. We therefore collect and label data from executed plans which we generate without a classifier. To do this, we run the planner without using \texttt{Classify} starting from some $\truestate_0$. Planning generates a transition sequence $\mathbb{T}^\reducedstate = \{T_t^\reducedstate | t=1,2,...\}$. We then execute the plan, which gives a ground-truth sequence of states $\tau^\truestate = \{\truestate_0, \truestate_t = \Gamma(\truestate_{t-1},\mathbb{T}^b_t.u^\reducedstate,\obstacle) | t=1,2,3...|\mathbb{T}^\reducedstate|\}$. We then reduce $\tau^\truestate$ to the reduced state space producing $\tau^\reducedstate = \{\tilde{\reducedstate}_t=r(\tau^\truestate_t,\obstacle)|t=1,2,...|\tau^\truestate|\}$. For time $t$, let the \textit{reduced dynamics prediction} be $\hat{\reducedstate}_t = \mathbb{T}^\reducedstate_t.\reducedstate$ and the \textit{rollout result} be $\tilde{\reducedstate}_t = \tau^b_t$. Figure \ref{fig:flowgraph} summarizes the computation required to produce these variables.

To label the data we require a function that evaluates if the two predictions are meaningfully similar for the given system. Let the function $\texttt{Close}:\mathcal{B} \times \mathcal{B} \rightarrow \{0,1\}$ return $1$ when two reduced states are meaningfully similar and $0$ otherwise. The environment $\obstacle$ is also an input to \texttt{Close} but we omit it for brevity. We label the $t$th transition in $\mathbb{T}^\reducedstate$ using the function $l$:
\begin{equation}
    \begin{aligned}
    l(t,\mathbb{T}^\reducedstate,\tau^\reducedstate) &= \begin{cases} 
        \texttt{Reliable}   &  \begin{aligned}\textbf{if } & \texttt{Close}(\hat{\reducedstate}_t, \tilde{\reducedstate}_t) \textbf{ and} \\ 
                                                           & \texttt{Close}(\hat{\reducedstate}_{t+1}, \tilde{\reducedstate}_{t+1})\end{aligned} \\
        \texttt{Unreliable} & \mbox{otherwise} \end{cases}
    \end{aligned}
    \label{eqn:labeling}
\end{equation}

Intuitively, this function first checks if $\hat{\reducedstate}_t$ $\tilde{\reducedstate}_t$ are close. If they are, and $\hat{\reducedstate}_{t+1}$ and $\tilde{\reducedstate}_{t+1}$ are not close, then the transition is labeled \texttt{Unreliable} because the reduced dynamics prediction was inaccurate. If the starting states $\hat{\reducedstate}_t$ and $\tilde{\reducedstate}_t$ \textit{are not} similar, then the rollout and reduced dynamics predictions have already diverged and we do not have a meaningful ground-truth label for this transition. To be conservative in our prediction, we label this transition \texttt{Unreliable}. If both $\hat{\reducedstate}_t$, $\tilde{\reducedstate}_t$ are close and $\hat{\reducedstate}_{t+1}$, $\tilde{\reducedstate}_{t+1}$ are close then $r$ and $g$ have performed well and we label the transition \texttt{Reliable}.

\vspace{-0.04in} 
\subsection{An Illustrative Navigation Example}
\vspace{-0.04in} 
\label{sec:car_example}
To clarify the above framework and learning problem formulation, we describe an example system where the various functions and spaces can be easily visualized. Consider a car with state $\truestate = [q,\dot{q}]$, where $q = [q_x,q_y,q_\theta]$ and control inputs $u^{\truestate} = [u^{\truestate}_a,u^{\truestate}_\phi]$, which correspond to the acceleration and steering angle. $f(x_t,\statecommand,\obstacle)$ is the standard second-order car dynamics. 

We define a reduced state using the function $\reducedstate = r(\truestate,\obstacle) = x_q$ (i.e. we only consider position variables in the reduced space) and the controls to be $u^\reducedstate = [\Delta x,\Delta y, \Delta \theta ]$. The reduced dynamics are $b_{t+1} = g(b_t,u^\reducedstate,\obstacle) = b_t + u^\reducedstate$.

Let the rollout function $x_{t+1} = \Gamma(x_t,u^\reducedstate,\obstacle)$ be a method that uses a controller to drive the car toward $r(x_t,E) + u^\reducedstate$. $\Gamma$ also checks if the car reaches the boundary of an obstacle in $\obstacle$ and will return the state on the boundary if it does. $\texttt{Close}(\hat{\reducedstate}_1,\tilde{\reducedstate}_1)$ outputs $1$ if the two reduced states are within a small Euclidean distance and $0$ otherwise. 

The task for the car is to drive to a given location while maintaining low speed and not colliding with obstacles. If we gather training data from this task domain we will find that when $u^\reducedstate$ drives the car toward an obstacle that is nearby, depending on the velocity at $x_0$, the car can hit the obstacle even though the lines between the planned $b_t$ and $b_{t+1}$ are collision-free for all $t$. Using only the planner, we would accept \textit{all} transitions where the line from $b_t$ to $b_{t+1}$ is collision-free. However, the classifier will learn that it is better to avoid transitions that entail driving past nearby obstacles. Using the classifier's output to further prune transitions will restrict the planner to transitions that are more likely to succeed when executing $\Gamma$ (see Fig. \ref{fig:classifier}).

\vspace{-0.04in} 
\subsection{What can be learned}
\begin{figure}[t]
    \centering
    \includegraphics[width=0.5\linewidth]{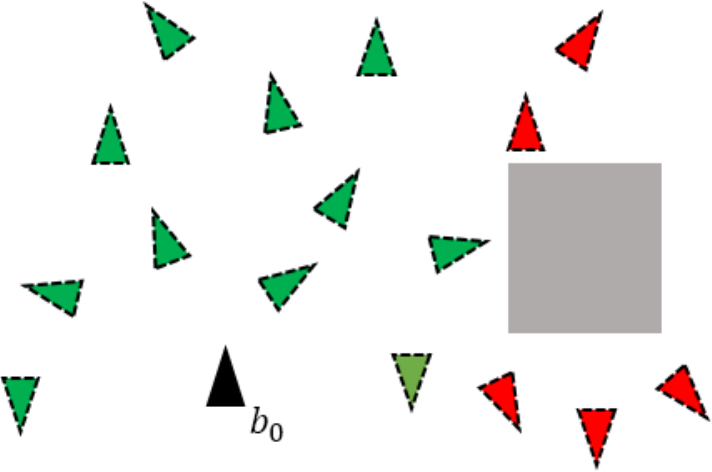}
    \vspace{-0.05in}
    \caption{Illustration of desired prediction from a classifier. Dotted triangles indicate $\hat{\reducedstate}_1$s from different $u^b_0$ inputs. Green: Classifier predicts these transitions are $\texttt{Reliable}$. Red: Classifier predicts these transitions are $\texttt{Unreliable}$. Note that the line between $b_0$ and $\hat{\reducedstate}_1$ is collision-free for all examples shown.}
    \label{fig:classifier}
    \vspace{0.1in}
    \centering
    \includegraphics[width=\linewidth]{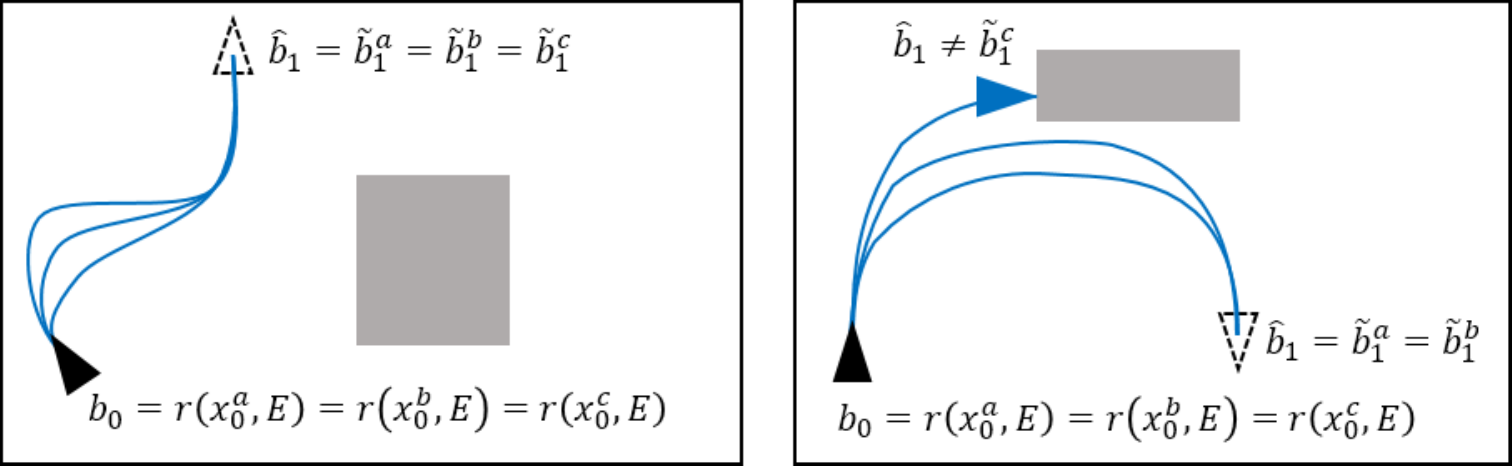}
    \caption{Illustration of the effect of information loss on the predictability of a transition. In both scenarios states with different velocities reduce to the same $b_0$. Left: A case where rolling out the same $u^b$ from different initial velocities (blue) produces the same $\tilde{b}_1$ values, since the controller is robust to initial velocity in this case. Right: A case where rolling out the same $u^b$ with different initial velocities produces different $\tilde{b}_1$ values. At high initial velocity ($c$) the controller cannot turn before reaching the obstacle.}
    \label{fig:cars}
\end{figure}

\vspace{-0.04in} 
While we may produce a useful classifier for planning, it is important to know that there is a fundamental limitation on what can be learned by this approach because of the loss of information that may happen in $r$ and/or $g$. Consider the following scenario: Let $b_0 = r(x_0^a,\obstacle) = r(x_0^b,\obstacle) = r(x_0^c,\obstacle)$, e.g. there are multiple states where the car is at a certain position but with a different velocity. If we apply reduced dynamics prediction for some $u^\reducedstate$, we obtain $\hat{\reducedstate}_1 = g(b_0,u^\reducedstate,\obstacle)$. However, we if do rollouts we obtain three resulting states: $x_1^a = \Gamma(x_0^a,u^\reducedstate,\obstacle), x_1^b = \Gamma(x_0^b,u^\reducedstate,\obstacle), x_1^c = \Gamma(x_0^c,u^\reducedstate,\obstacle)$, and then three reduced states: $\tilde{\reducedstate}_1^a = r(x_1^a,\obstacle),\tilde{\reducedstate}_1^\reducedstate = r(x_1^b,\obstacle), \tilde{\reducedstate}_1^c = r(x_1^c,\obstacle)$. It may be the case that $\texttt{Close}(\hat{\reducedstate}_1, \tilde{\reducedstate}_1^k)$ does not produce the same result for $k={a,b,c}$ (an example for the car system is shown in Figure \ref{fig:cars}). In terms of classification, this is a case of noisy labeling, and many methods have been devised to address this problem (e.g. SVM with slack variables). 

However, if there are many noisy labels in the data, it means that $r$ and $g$ are not useful for this task domain. As a result the classifier will not make meaningful predictions and we would expect that it would provide no benefit over simply planning in $\reducedstatespace$. However, in our experiments with a planar arm and with rope manipulation we found that we do indeed see a benefit when using the classifier.

\vspace{-0.04in} 
\subsection{Using the Classifier in Planning}
\vspace{-0.04in} 

While it is possible to query every transition considered by the planner and reject all those that are classified as $\texttt{Unreliable}$, such a strategy would likely be overly optimistic about what the classifier has learned. In difficult scenarios the classifier may erroneously reject a set of transitions which is necessary to reach the goal, thus causing the planner to fail. Thus we accept transitions that are classified as \texttt{Unreliable} with a small probability determined by parameters $k$, a manually-specified constant, and $p_{acc}$, the validation accuracy of the classifier (see  Algorithm~\ref{alg:checktransition}). $p_{acc}$ is included because we wish to be more permissive about accepting transitions when the classifier performs more poorly in terms of generalization.

While the above approach of incorporating classification can be applied to a wide range of planners, in this paper we focus on using RRT-based planners. An advantage of this approach for RRT-based planners is that we maintain the probabilistic completeness properties of the planner by guaranteeing that any transition will be accepted with a non-zero probability (although the probability is small for transitions classified as $\texttt{Unreliable}$).


\vspace{-0.04in} 
\section{APPLICATION TO A TORQUE-LIMITED PLANAR ARM}
\vspace{-0.04in} 

First, we demonstrate our framework on a 3-link arm that moves in the X-Z plane with gravity in the $-z$ direction. For this system we focus on the effects of including a classifier in the planner without using a reduction function. This allows us to disentangle the effects of model reduction and inaccurate dynamics. 

\begin{algorithm}[t]
\caption{\strut \texttt{CheckTransition}($T^b$)}
\begin{algorithmic}[1]
    \algtext*{EndIf}
    \State $\reducedstate' \gets g(T^\reducedstate.\reducedstate,  T^\reducedstate.\reducedcommand, \obstacle)$
    \If {\texttt{Classify}$(T^\reducedstate) ==$ \texttt{Reliable}}
        \State \Return $\reducedstate'$
    \EndIf
    \State $a \sim U[0, 1]$
    \If {$a < e^{-k p_{\text{acc}}}$}
        \State \Return $\reducedstate'$
    \EndIf
    \State \Return \texttt{$\emptyset$}
\end{algorithmic}
\label{alg:checktransition}
\end{algorithm}

For this experiment we use the MuJoCo simulator~\cite{mujoco} as the ground truth dynamics. Each joint is controlled using the default position servo actuator available in MuJoCo.
We convert the MuJoCo simulation to a quasistatic system by waiting for the arm to settle after a configuration is commanded (this is $f$).
These experiments do not have any obstacles, so we omit $\obstacle$ for brevity.

\vspace{-0.04in} 
\subsection{Problem Statement}
\vspace{-0.04in} 
Let $\truestate \in \reals^3$ be the true state of the system. Let $u^\truestate = \truestate_{des} \in \reals^3$ be the commanded position of the arm at each timestep. Let $f(x_t, u_t)$ be the quasistatic dynamics defined by MuJoCo. We set torque limits $\tau_1, \tau_2, \tau_3$ so that $\tau_1 \ll \tau_2 = \tau_3$. This means the first joint cannot support the weight of the arm when extended horizontally. As we are not doing a model reduction, $\reducedstate = r(\truestate) = \truestate$. Commands in both spaces are also the same: $u^\reducedstate = u^\truestate$. The dynamics in $\reducedstatespace$ are purely kinematic: $\reducedstate_{t+1} = g(\reducedstate_t, u^\reducedstate_t) = u^\reducedstate_t$. These dynamics are fast to evaluate and therefore efficient for planning, but can result in plans which do not reach the goal configuration when executed (Fig.~\ref{fig:planar_arm}). As there are no obstacles and $u^\reducedstate = u^\truestate$, the rollout function $\Gamma(x_t, u^\reducedstate_t, \obstacle) = f(x_t, u_t^\truestate)$.

The planning problem is for the arm to reach a goal end-effector position. Using the true dynamics, this corresponds to Problem \eqref{eqn:main_planning_problem}. However, since we assume the true dynamics are not available, we seek to solve Problem \eqref{eqn:approx_planning_problem} given the definitions of $r$ and $g$ above.

\vspace{-0.04in} 
\subsection{Data Collection, Labelling, and Training}
\vspace{-0.04in} 
To collect training data we initialized the system from random start configurations, planning to random goal configurations using RRT-Connect~\cite{kuffner2000rrt}. In practice these are straight lines in configuration space after smoothing the path. This generated a total of 231,815 transitions to use in training and validation. A randomly selected 20\% of the data is held out for validation. We define $\texttt{Close}(\hat{\reducedstate}_t, \tilde{\reducedstate}_t)$ based on the Euclidean distance between configurations:
\begin{equation}
    \texttt{Close}(\hat{\reducedstate}_t, \tilde{\reducedstate}_t) = \| \hat{\reducedstate}_t - \tilde{\reducedstate}_t \| < \alpha
\end{equation}
with $\alpha = 0.075$.

For this planar arm, our classifier is a neural network that takes $\reducedstate$ and $\reducedstate'$ as input. The network has three hidden layers with 256, 128, and 64 hidden units respectively and a single output neuron. The hidden units use a Leaky ReLu activation~\cite{maas2013rectifier}, and the output uses a sigmoid activation. With this network we achieve 99\% training accuracy and 99\% validation accuracy within 4 epochs.

\vspace{-0.04in} 
\subsection{Planning and Results} 
\vspace{-0.04in} 

\begin{figure}[t]
    \centering
    \includegraphics[width=\columnwidth]{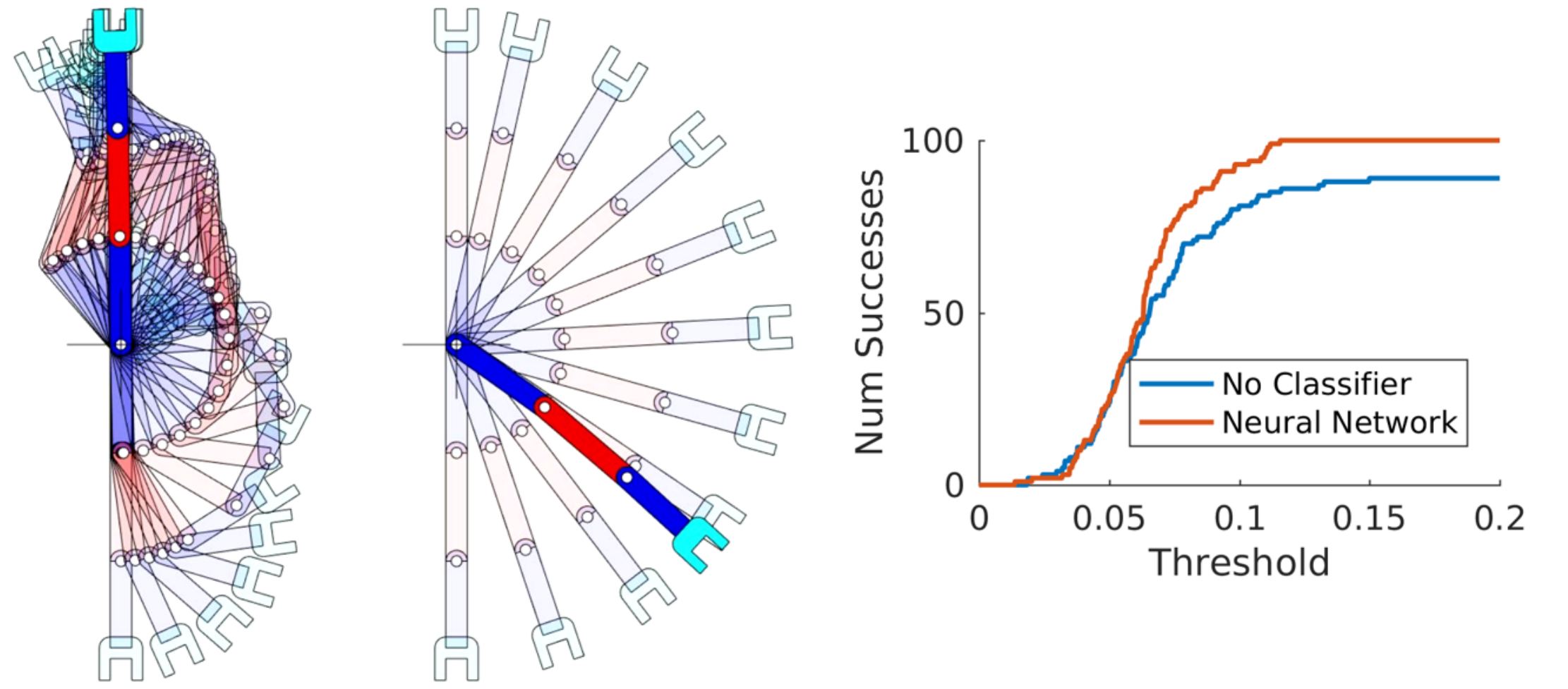}
    \vspace{-0.1in}
    \caption{Left: Plan generated using the learned classifier to go from $[\frac{-pi}{2}, 0, 0]$ to $[\frac{pi}{2}, 0, 0]$. The plan avoids transitions which move the arm toward a horizontal position and successfully completes the task. Center: Plan generated without the classifier. The plan takes the arm to the horizontal position where it fails due to the torque limit. Right: Number of successes as success threshold $\beta$ varies}
    \label{fig:planar_arm}
\end{figure}

To test the effect of using the learned classifier, we randomly generate 100 planning queries and evaluate how well generated plans perform when executed. Each planning query consists of a random start configuration in $\reals^3$ and goal IK solutions for a random target point in $\reals^2$. We only consider queries for which there is at least one IK solution where the controller can maintain the configuration of the arm within $\alpha = 0.075$ of the IK solution. For planning we use the OMPL~\cite{ompl} implementation of RRT-Connect, setting the probability scale factor $k$ to 1, and $p_{acc}$ to 0.99. The resulting path is post-processed using the default \texttt{simplifyMax} options. A plan is considered successful if the distance between the final configuration and any IK solution is less than a threshold $\beta$ (Fig.~\ref{fig:planar_arm}). Tests are performed on an i5-3570K @ 4.3 GHz. Planning and simplification takes approximately 2 ms without a classifier, and approximately 7 seconds with our classifier. For small $\beta$, using the classifier does not improve performance due to the steady state error inherent in the system. As $\beta$ increases, we see that the planner that uses a classifier is able to successfully find paths to all queries while the baseline is unable to succeed at some queries (see attached video).

\vspace{-0.04in} 
\section{APPLICATION TO ROPE MANIPULATION}
\vspace{-0.04in} 

We now present the application of our framework to rope manipulation, where we use both a reduction of the state space and an approximate dynamics model.

\vspace{-0.04in} 
\subsection{Problem Statement}
\vspace{-0.04in} 

Let the robot be represented by a pair of grippers with configuration $\robotconfig \in \robotconfigspace$. We assume that the robot configuration can be measured exactly. In this work we assume the robot to be a set of free floating grippers; in practice we can track the motion of these with inverse kinematics.

We assume that our model of the robot is purely kinematic, with no higher order dynamics. We assume that the robot has two end-effectors that are rigidly attached to the rope. The configuration of a rope is a set $\deformconfig \subset \reals^3$ of $\numdeformpoints = | \deformconfig |$ points. We assume that we have a method of sensing $\deformconfig$. The rest of the environment $\obstacle$ is assumed to be both static, and known exactly. We assume that the robot moves slowly enough that we can treat the combined robot and rope as quasi-static. The true state of the system is then $\truestate = [\robotconfig,\deformconfig]$ and $\statecommand = \Delta \robotconfig$. Let $f$ be a joint-space controller for the robot that stops when any of the following occur: 1) the grippers contact an object; or 2) the object stretches by more than a factor $\lambda$. Due to the difficulty of simulating rope physics, we do not assume we can execute $f$ within a motion planner.

We wish to find a sequence of $\taskexecutiontime$ commands $\{\statecommand_1, \dots, \statecommand_{\taskexecutiontime} \}$ to move from a start state $\truestate_0$ to a goal gripper configuration $\robotconfig_g$ such that each motion is feasible (this corresponds to Problem \eqref{eqn:main_planning_problem}). Note that this planning problem does not require bringing the rope to a specific configuration, which can often be done using local control \textit{after} bringing the object to a desired area (as in~\cite{McConachie2019}). 
Because we do not have access to $f$, we cannot solve this problem by planning in $\truestatespace$ directly.




To make planning tractable we will perform a reduction and learn a classifier from data to predict when the reduction can be trusted. That classifier will then be used in a motion planner to bias it away from transitions that are not likely to be feasible under $\Gamma$.

\vspace{-0.04in} 
\subsection{Reduction}
\vspace{-0.04in} 
\label{sec:isrr_model_approx}

\cite{McConachie2017} introduced the idea of a \textit{virtual elastic band} (VEB) between the robot's end-effectors. This VEB represents the shortest path through the rope between the end-effectors. The band approximates the constraint imposed by the rope on the motion of the robot; if the end-effectors move too far apart, then the VEB will be too long, and thus the rope is stretched beyond a task-specified maximum stretching factor. Similarly, if the VEB gets caught on an obstacle and becomes too long, then the rope is also overstretched. By considering only the geodesic between the end-effectors, we are assuming that the rest of the rope will comply to the environment, and does not need to be considered when predicting overstretch. The VEB representation allows us to use a fast prediction method when planning, but does not account for the part of the material that is slack. Denote the configuration of a VEB (i.e. a sequence of points) as $\band$. Then $\reducedstate = [\robotconfig,\band]$ and can be generated by $\reducedstate = r(\truestate,\obstacle)$ for the reduction function defined in Section 4.2.2 of~\cite{McConachie2017}. The dynamics of a VEB, $g(\reducedstate,\reducedcommand,\obstacle)$, are based on Quinlan's path deformation algorithm as presented in \cite{Quinlan1994} (see Section~4.2.3 of \cite{McConachie2017}). We also augment $g$ to return $\emptyset$ when $\reducedcommand$ causes the object to become overstretched. 
$b$ is then propagated using $g$ and $\reducedcommand = \statecommand$. Since the commands are the same, our rollout function $\Gamma$, which uses $\reducedcommand$, is equivalent to $f$. To find a path in $\reducedstatespace$ we must solve Problem \eqref{eqn:approx_planning_problem}.




We use the planner described in \cite{McConachie2019} to solve this problem; this is an RRT-based planner designed for use with virtual elastic bands as part of the planning configuration space. This planner searches for a feasible path for the robot between a given start and goal configuration, while ensuring the VEB is never overstretched.

The virtual band approximation choice favors speed over model accuracy; as a consequence, there are several issues that it does not address. Specifically, environments with ``hooks'' can cause problems due to the approximation methods: The virtual elastic band assumes that there is no minimum length of the rope. This assumption means that the planner cannot detect cases where the slack material can get caught on corners or hooks, preventing the motion plan from being completed because the caught object can overstretch. To reduce the chances of this occurring we learn a classifier for $T^\reducedstate$ to predict if a given transition is either \texttt{Reliable} or \texttt{Unreliable} and bias the planner away from \texttt{Unreliable} transitions. We bias the planner using the \texttt{CheckTransition} function shown in Algorithm~\ref{alg:checktransition}, which is used as an edge validity check in addition to the collision and overstretching checks described in \cite{McConachie2019}.

\vspace{-0.04in} 
\subsection{Learning the Classifier}
\vspace{-0.04in} 
To learn the classifier we define the \texttt{Close} function for our domain to be:
$$
\texttt{Close}(\hat \reducedstate_t, \tilde \reducedstate_t) = \left( D(\hat \reducedstate_t.\band, \tilde \reducedstate_t.\band) < \alpha \right) \land \texttt{FOH}(\hat \reducedstate_t.\band, \tilde \reducedstate_t.\band,\obstacle)
$$
\noindent Where $D$ computes the sum of the point-wise distances between the two virtual elastic bands, $\alpha$ is a constant, and \texttt{FOH} evaluates if the two bands are in the same first-order homotopy class \cite{Jaillet2008}. We then apply labeling function $l$ shown in Eq. \ref{eqn:labeling} for each transition. We generate many plans using our planner (without the classifier) to produce the dataset (see Section \ref{sec:ropedata}).

\begin{figure}[t]
    \centering
    \includegraphics[width=0.4\textwidth,keepaspectratio]{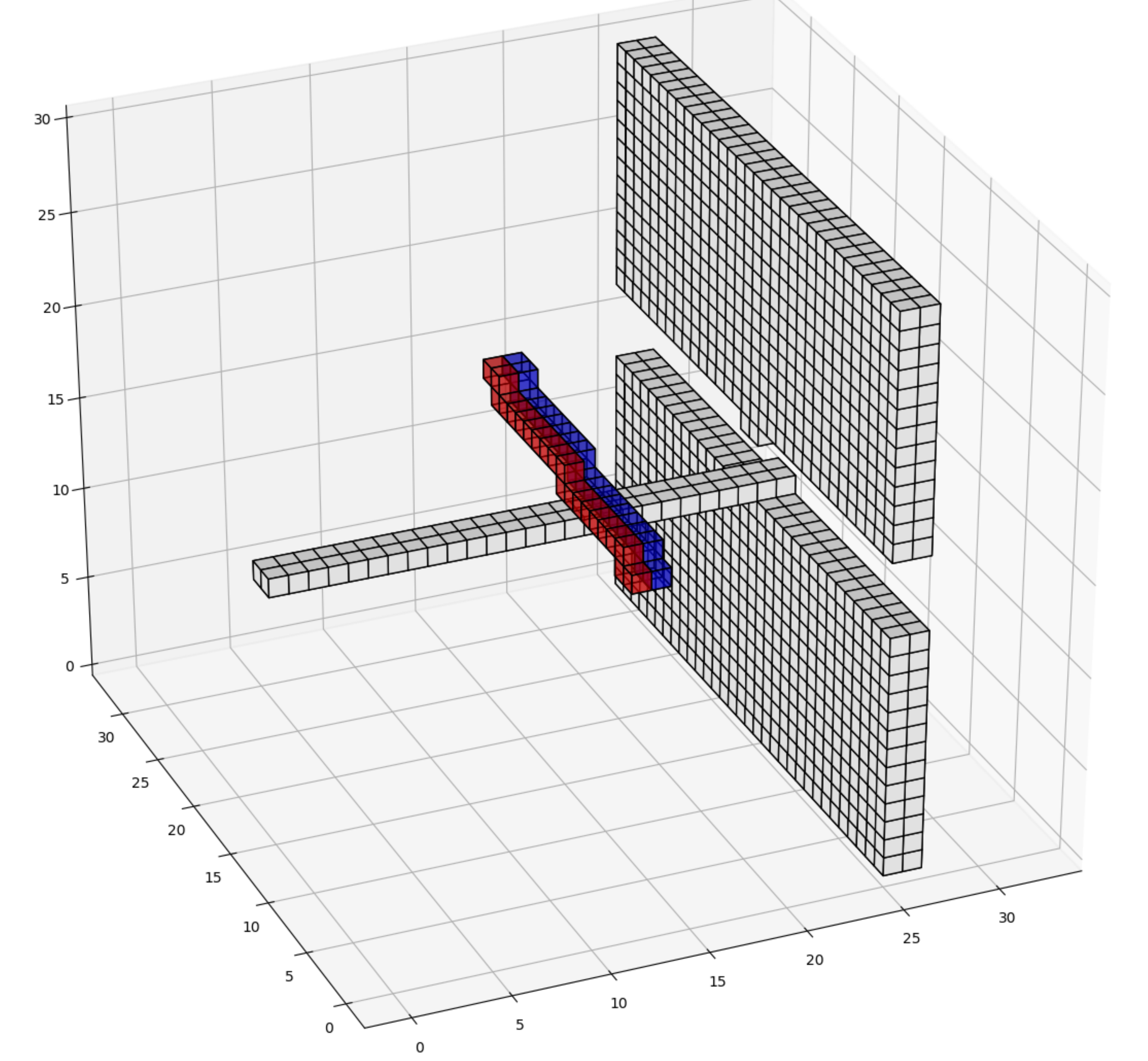}
    \caption{Input to the VoxNet classifier is a 3-channel voxel grid, where white is the local environment, red is the pre-transition band, and blue is the post transition band. Positions outside the bounds of the environment are marked as occupied in the local environment channel.}
    \vspace{-0.1in}
    \label{fig:VoxNet_input}
\end{figure}


For our classifier we use a neural network based on \textit{VoxNet} \cite{Maturana2015VoxNet}, a network for classifying objects from a voxel grid. The network consists of two 3D convolutional layers (filter size 5, 3 and stride 2, 1 respectively) with max pooling followed by two fully-connected layers. All layers have ReLU activations except the output layer, which has a sigmoid activation.

The input for our classifier consists is a three-channel binary voxel grid, with channels  $\big<\obstacle, b, b'\big>$, where $b' = g(b, u^b, E)$. Each voxel grid is $32\times32\times32$, and is constructed by checking for occupancy at every cell center. An example of the voxelized representation is shown in Fig.~\ref{fig:VoxNet_input}.


\begin{figure*}[t]
    \vspace{-0.02in}
    \centering
    \includegraphics[height=1.8in,keepaspectratio]{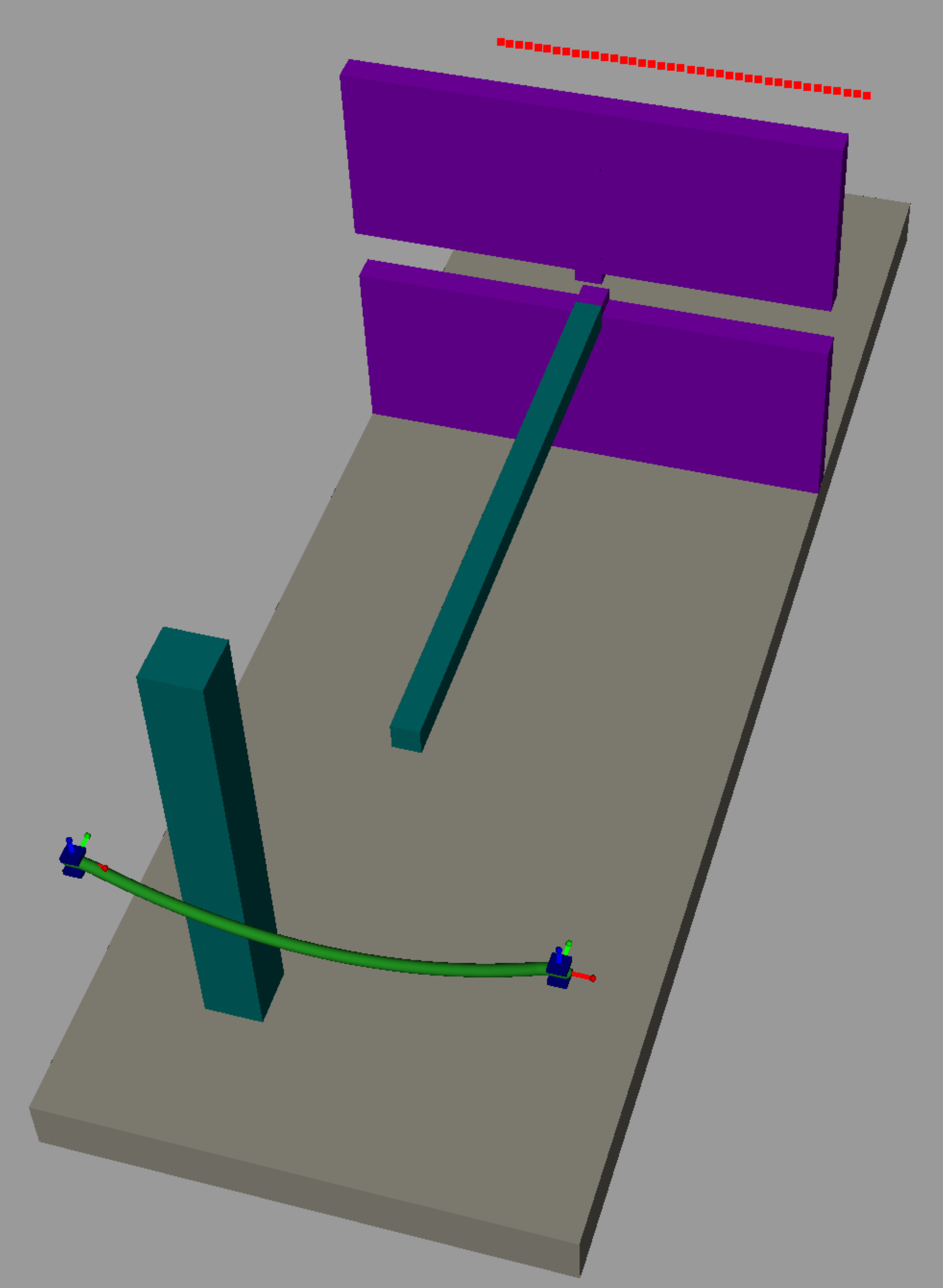}\hfill
    \includegraphics[height=1.8in,keepaspectratio]{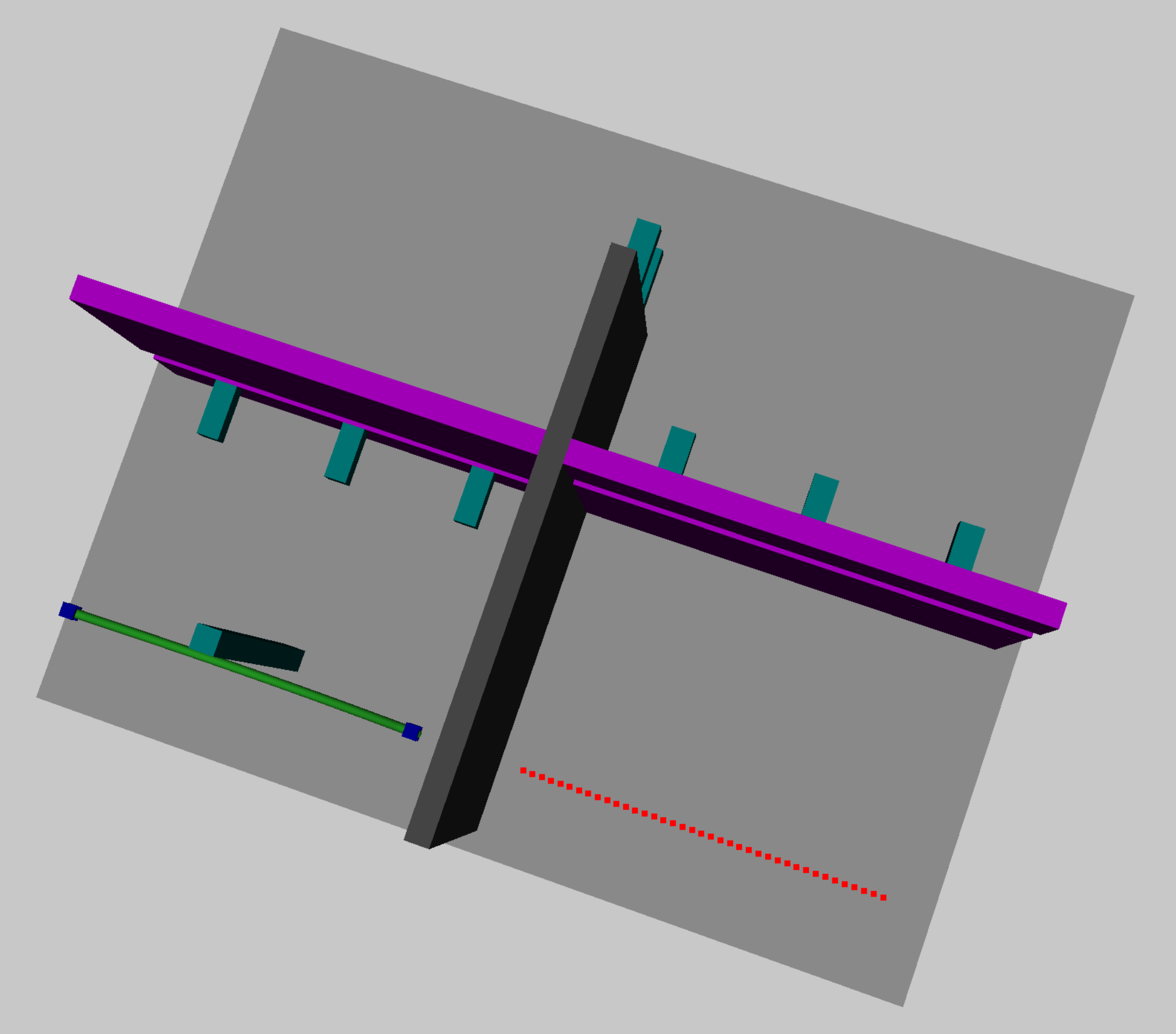}\hfill
    \includegraphics[height=1.8in,keepaspectratio]{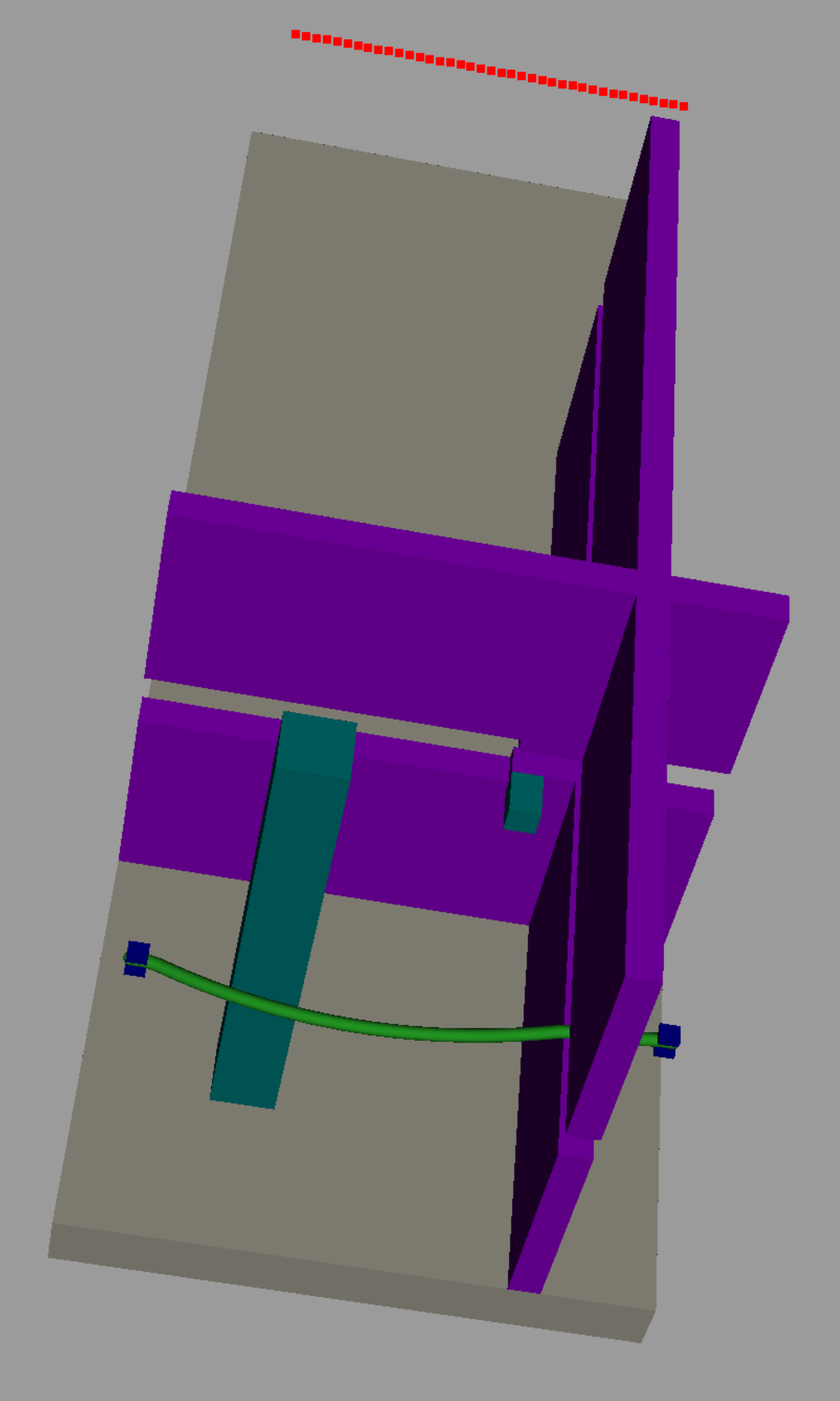}\hfill
    \includegraphics[height=1.8in,keepaspectratio]{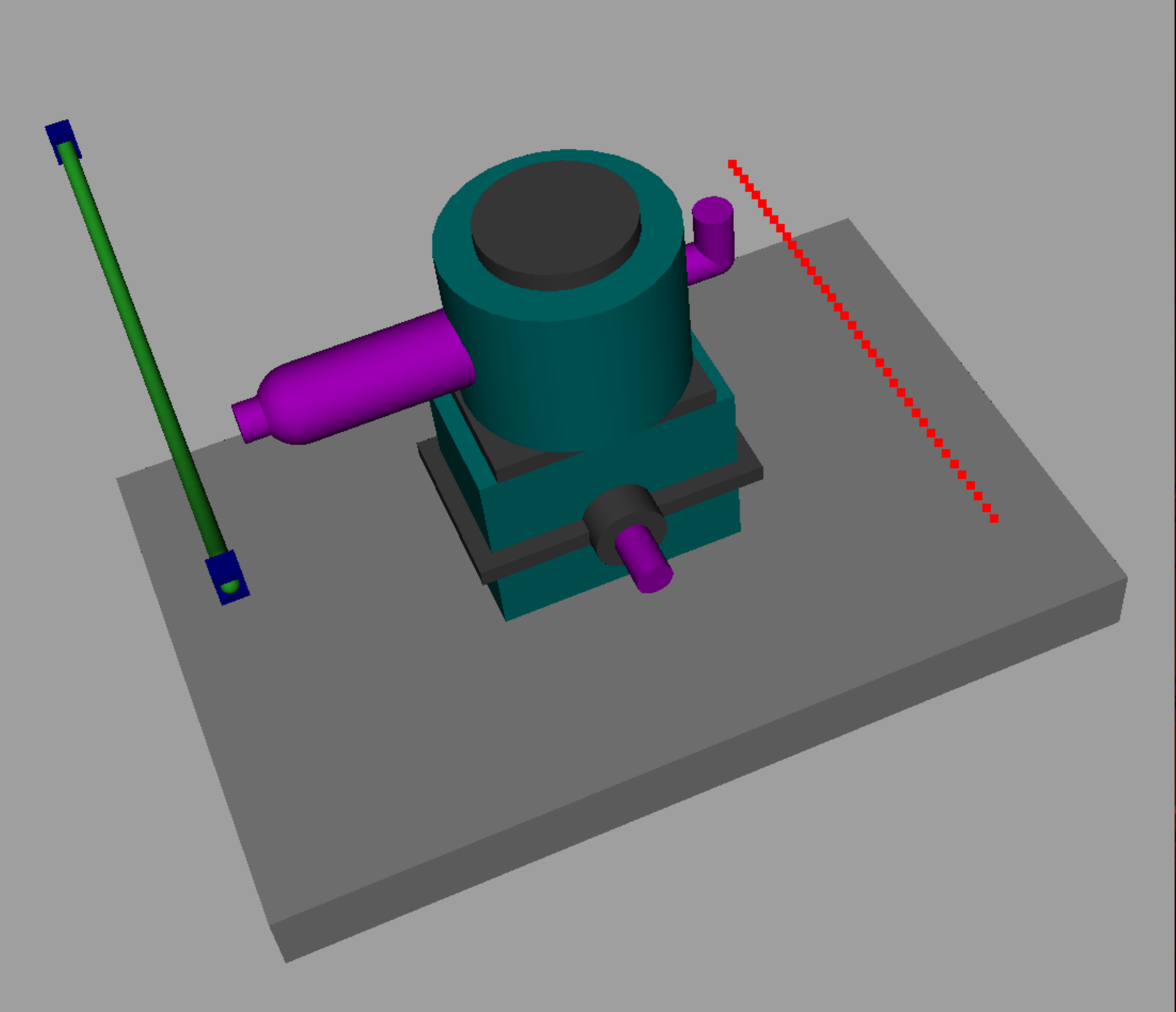}%
    \vspace{-0.1in}
    \caption{The rope is shown in green, with the grippers shown in blue. The target area for the grippers is shown in red. Walls with narrow slits for the grippers are shown in purple. Hooks and other obstacles are shown in dark cyan. Left: Simple Hook; Center Left: Multi Hook; Center Right: Complex Hook; Right: Engine Assembly}
    \label{fig:scenarios}
    \vspace{-0.25in}
\end{figure*}


\vspace{-0.04in} 
\section{ROPE MANIPULATION EXPERIMENTS}
\vspace{-0.04in} 

To characterize the planner performance with the classifier, we designed seven simulation scenarios where the rope must be moved from one side of the scene to the other, along with one physical experiment for real-world validation.
All simulation experiments were conducted in the open-source Bullet simulator \cite{Coumans2010}, with additional wrapper code developed at UC Berkeley \cite{ucberkley_bullet}. The rope is modeled as a series of small capsules linked together by springs. We emphasize that our method does not have access to the model of the rope or the simulation parameters. The simulator is used as a ``black box'' for testing. All tests are performed using an i7-7700K 4.2 GHz CPU with 32 GB of RAM. For all experiments we set $\lambda$ to $1.15$, and $\alpha$ to $0.5$.

\vspace{-0.04in} 
\subsection{Scenarios}
\vspace{-0.04in} 

Each scenario involves moving the rope past one (or more) hooks, while the grippers traverse a narrow slit (Fig.~\ref{fig:scenarios}).

\subsubsection{Simple Hook}

In the Simple Hook environment, the end of the hook is not near any obstacles, thus the planner does not need to deal with the end of the hook and the narrow slit at the same time. We test four variants of the simple hook environment: Short, Regular, Long, Very Long corresponding to the lengths of the rope which are 0.55m, 0.87m, 1.13m, and 1.59m respectively.

\subsubsection{Multi Hook}

In the Multi Hook environment, the rope must pass through three series of hooks, and two narrow slots before reaching the red region on the far side of a solid wall. The rope has length 0.87m.

\subsubsection{Complex Hook}

In the Complex Hook environment, the grippers are forced to pass on opposite sides of a small hook, while also passing through a narrow slit. The rope has length 0.87m.

\subsubsection{Engine Assembly}
In the Engine Assembly environment, we seek to move the grippers from one side of an engine model \cite{EngineAssemblyTumber} to the others avoiding two hooks on the front and back of the engine. The rope has length 0.87m.
 The engine assembly environment is shown in Fig. \ref{fig:scenarios}.

\subsubsection{Physical Robot}
In the Physical Robot environment, we attempt the engine assembly task on a physical 16 DoF robot shown in Fig. \ref{fig:intro_figure} with a 3D printed model of the engine and a rope of length 0.46m.

\vspace{-0.04in} 
\subsection{Data Collection}
\vspace{-0.04in} 
\label{sec:ropedata}
To collect training data, we ran the planner without any classifier on the Simple Hook Regular scene repeatedly, generating a total of 4190 plans from many different starting locations. This generated a total of 562,177 transitions to use in training and validation. This training set is generated only from the Simple Hook environment using the Regular length rope. We emphasize that we use the classifier trained on this data for planning in \textit{all} test environments. 

\vspace{-0.04in} 
\subsection{Training the Classifier}
\vspace{-0.04in} 
VoxNet is trained using the Adam optimizer~\cite{adamOptimizer2015} with initial learning rate of $5 \times 10^{-4}$. We use a learning rate decay of $0.8$ every 4 epochs. Since the dataset set is imbalanced, with 32\% of the examples labelled as unreliable, and 68\% labelled as reliable, we use a weighted random sampler to make all minibatches balanced. A randomly selected 10\% of the data is held out for validation. Our minibatch size is 32, and we train for 100 epochs. We use the binary cross-entropy loss function during training. Training took approximately 16 hours on a Tesla V100-SMX2 GPU. The VoxNet classifier achieved 99\% accuracy on the training set and 91\% accuracy on the validation set.

\vspace{-0.04in} 
\subsection{Planning Results}
\vspace{-0.04in} 

\begin{table}[t]
\begin{center}
\begin{tabular}{|l|l|c|c|}
\hline
\multirow{2}{*}{Environment} & \multirow{2}{*}{Metric} & \multicolumn{2}{c|}{Classifier} \\ 
\cline{3-4}
& &\begin{tabular}[c]{@{}c@{}}None \end{tabular} & \begin{tabular}[c]{@{}c@{}}VoxNet\end{tabular} \\
\hline 
\multirow{3}{*}{Simple Hook - Short}        & Success rate          & 18/30 & \textbf{30/30} \\
                                            & Planning time (s)     & 0.6   & 14.6 \\
                                            & Smoothing time (s)    & 1.0   & 5.4 \\
\hline
\multirow{3}{*}{Simple Hook - Regular}      & Success rate          & 20/30 & \textbf{29/30} \\
                                            & Planning time (s)     & 3.7   & 17.3 \\
                                            & Smoothing time (s)    & 4.0   & 6.5 \\
\hline
\multirow{3}{*}{Simple Hook - Long}         & Success rate          & 23/30 & \textbf{29/30} \\
                                            & Planning time (s)     & 6.8   & 51.9 \\
                                            & Smoothing time (s)    & 6.4   & 8.0 \\
\hline
\multirow{3}{*}{Simple Hook - Very Long}    & Success rate          & 18/30 & \textbf{27/30} \\
                                            & Planning time (s)     & 12.4  & 15.4 \\
                                            & Smoothing time (s)    & 15.6  & 11.4 \\
\hline
\multirow{3}{*}{Multi Hook}                 & Success rate          & 9/30  & \textbf{13/30} \\
                                            & Planning time (s)     & 11.5  & 44.1 \\
                                            & Smoothing time (s)    & 22.0  & 30.0 \\
\hline
\multirow{3}{*}{Complex Hook}               & Success rate          & 0/30  & \textbf{20/30} \\
                                            & Planning time (s)     &  3.7  & 28.7 \\
                                            & Smoothing time (s)    & 27.0  & 23.4 \\
\hline
\multirow{3}{*}{Engine}                     & Success rate          & 1/30  & \textbf{10/30} \\
                                            & Planning time (s)     & 4.8   & 2.0 \\
                                            & Smoothing time (s)    & 0.3   & 4.1 \\
\hline
\end{tabular}
\end{center}
\caption{Planning statistics, averaged over 30  trials}
\label{tab:planning_stats}
\end{table}

To evaluate the planning performance when using the classifier, we generated 30 plans using the classifier on each test environment and compare the success rate and planning time to planning without a classifier. A success is when executing the plan results in a final gripper configuration which is within a small tolerance of the goal gripper configuration. If, for example, the rope gets caught on a hook and prevents the grippers from reaching the goal, the trial is marked as a failure. Results are shown in Table~\ref{tab:planning_stats}. We set $k$ to 10 and $p_{acc}$ to $0.91$. Our results show that using a classifier improves the success rate of the planner over not using a classifier in all tested scenarios, but the effect is less prominent on the Multi Hook environment. The use of a classifier does lead to longer planning time, partially due to extra computation when checking each edge, as well as making the planning problem harder to solve. Sec.~\ref{sec:discussion} discusses this in more detail. Example results can be found in the attached video.


\vspace{-0.04in} 
\section{DISCUSSION}
\label{sec:discussion}
\vspace{-0.04in} 

We found that the use of a neural network classifier to evaluate the reliability of a model approximation can be an effective way to improve the success rate of a planner for rope manipulation. When using the classifier in the planner as a hard constraint ($k = \infty$), some starting configurations would cause the classifier to reject all transitions from that state, leading to planning failure. An interesting approach to setting $k$ would be to treat it similar to temperature as done in T-RRT~\cite{Jaillet2008transition}. Despite training the rope manipulation classifier on only a single rope and environment (the Simple Hook with Regular length rope), we found that the classifier was able to generalize and lead to improved planning performance with both different lengths of rope and more complex environments. 

It is important to note that while our method can find more feasible plans than not using a classifier, we may be making the planning problem more difficult. In the planar arm example, a straight line in configuration space between start and goal solves the planning problem under $g$ but may not be a feasible plan under $\Gamma$. To avoid this mismatch, the classifier restricts the set of transitions that can be used, but doing so may induce a narrow passage effect, which leads to longer planning times.

A major challenge in this work is determining how to label the data in a way that will lead to good performance. In particular, by including transitions where the reduced dynamics predictions have already diverged in our dataset and labelling these transitions as \texttt{Unreliable}, the classifier learns that interaction with objects is frequently poorly modelled by the VEB. This leads the planner to avoid contact with the environment when possible, but many interesting tasks involving deformable objects will explicitly require interaction with other objects.

One interesting point is that increased classification accuracy does not necessarily lead to better planning performance. We experimented with different classifiers during development, and although VoxNet had the best classification accuracy on the validation set and usually the best performance, other classifiers performed equivalently or better for planning in some environments. It would therefore be desirable to find a way to use planner success to train the classifier, and we plan to investigate this in future work.


\vspace{-0.04in} 
\section{CONCLUSION}
\vspace{-0.04in} 
\label{sec:conclusion}

This paper proposed a novel formulation of planning in reduced state spaces that directly addresses the challenge of generating plans that are feasible in the true state space while planning in a reduced space with approximate dynamics. We addressed this model mismatch problem by introducing the use of a classifier into the planning process. This classifier is trained to classify state transitions in the reduced state space as reliable or unreliable. Our experiments on both a torque-limited 3-link planar arm and rope manipulation tasks show that 1) we can learn a classifier from training data with high validation accuracy; and 2) using this classifier to bias the planner away from unreliable transitions improves success rate in all tested scenarios.

\vspace{-0.04in} 
\bibliographystyle{IEEEtran}
\bibliography{references_strings_abbrev,references_dale}

\end{document}